\documentclass[10pt,twocolumn,letterpaper]{article}

\usepackage{cvpr}
\usepackage{times}
\usepackage{epsfig}
\usepackage{graphicx}
\usepackage{amsmath}
\usepackage{amssymb}
\usepackage{adjustbox}
\usepackage{array}
\usepackage{multirow}
\usepackage{comment}
\usepackage{authblk}
\newcommand*{\email}[1]{\texttt{#1}}
\newcommand*{\affaddr}[1]{#1} 
\newcommand*{\affmark}[1][*]{\textsuperscript{#1}}

\newcommand{\rpm}{\raisebox{.2ex}{$\scriptstyle\pm$}}
\newcommand{\rpmaccurate}{\sbox0{$1$}\sbox2{$\scriptstyle\pm$}\raise\dimexpr(\ht0-\ht2)/2\relax\box2 }
\newcommand\blfootnote[1]{%
  \begingroup
  \renewcommand\thefootnote{}\footnote{#1}%
  \addtocounter{footnote}{-1}%
  \endgroup
}

\usepackage[pagebackref=true,breaklinks=true,letterpaper=true,colorlinks,bookmarks=false]{hyperref}

\cvprfinalcopy 
\usepackage{cleveref}
\def\cvprPaperID{5185} 

\title{Out-of-Distribution Detection for Generalized Zero-Shot Action Recognition}

\author{%
\vspace{-0.5cm}
Devraj Mandal*\affmark[1], Sanath Narayan*\affmark[2], Saikumar Dwivedi\affmark[3], Vikram Gupta\affmark[3], Shuaib Ahmed\affmark[3], \\ \vspace{-0.35cm}Fahad Shahbaz Khan\affmark[2], and Ling Shao\affmark[2]\\
\affaddr{\affmark[1]Indian Institute of Science, Bangalore} \quad
\affaddr{\affmark[2]Inception Institute of Artificial Intelligence, UAE}\\
\affaddr{\affmark[3]Mercedes-Benz R\&D India, Bangalore}\\
\email{\affmark[1]\small devrajm@iisc.ac.in}\quad 
\email{\affmark[2]\small firstname.lastname@inceptioniai.org} \\
\email{\affmark[3]\small firstname.lastname@daimler.com}
}

\begin{document}
\maketitle
\begin{abstract}
    
\blfootnote{$*$Authors contributed equally.}
\blfootnote{Code available at https://github.com/naraysa/gzsl-od}
Generalized zero-shot action recognition is a challenging problem, where the task is to recognize new action categories that are unavailable during the training stage, in addition to the seen action categories. Existing approaches suffer from the inherent bias of the learned classifier towards the seen action categories. As a consequence, unseen category samples are incorrectly classified as belonging to one of the seen action categories. In this paper, we set out to tackle this issue by arguing for a separate treatment of seen and unseen action categories in generalized zero-shot action recognition. We introduce an out-of-distribution detector that  determines whether the video features belong to a seen or unseen action category. To train our out-of-distribution detector, video features for unseen action categories are synthesized using generative adversarial networks trained on seen action category features. To the best of our knowledge, we are the first to propose an out-of-distribution detector based GZSL framework for action recognition in videos. Experiments are performed on three action recognition datasets: Olympic Sports, HMDB51 and UCF101. For generalized zero-shot action recognition, our proposed approach outperforms the baseline~\cite{Xian2018} with absolute gains (in classification accuracy) of 7.0\%, 3.4\%, and 4.9\%, respectively, on these datasets.
\end{abstract}

\section{Introduction}
Zero-shot learning (ZSL) is a challenging problem, where the task is to classify images or videos into new categories that are unavailable during the training stage. 
Generalized zero-shot learning (GZSL), introduced in ~\cite{zsl-good-bad-ugly}, differs from ZSL in that the test samples can belong to the seen or unseen categories. The task of GZSL is therefore harder than ZSL due to the inherent bias of the learned classifier towards the seen categories. In this paper, we focus on the problem of generalized zero-shot action recognition in videos and treat ZSL as a special case of GZSL.

Most existing approaches ~\cite{hmdb,karpathy14sports1m,tran15c3d,carreira17i3d} tackle the problem of action recognition in videos in a fully-supervised setting. In such a setting, all the action categories that occur during testing are known \emph{a priori}, and instances from all action categories are available during training. However, the fully-supervised problem setting is unrealistic for many real-world applications (\eg, automatic tagging of actions in web videos), where information regarding some action categories is not available during training. Therefore, in this work we tackle the problem of action recognition under zero-shot settings.

\begin{figure}[t]
    \centering
    \includegraphics[width=0.47\textwidth,height=0.3\textwidth]{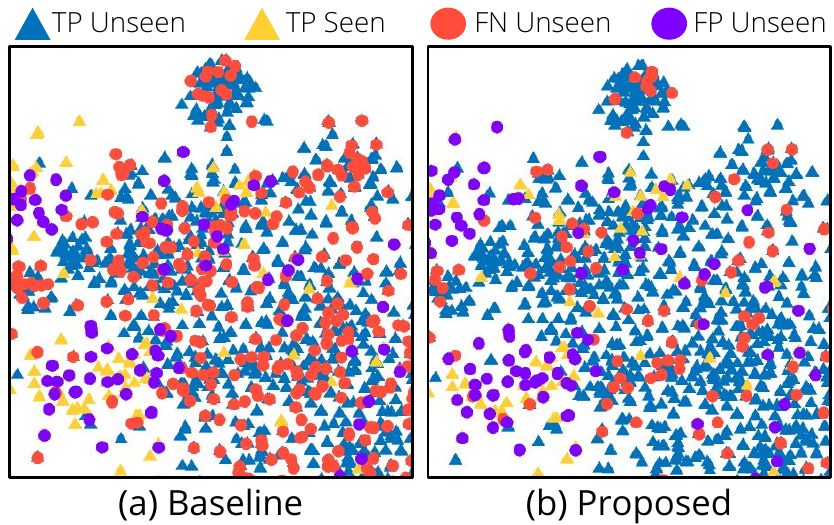}
    \caption{Illustration of the bias reduction achieved by the proposed framework on a random test split of the HMDB51 dataset. On the left: t-SNE scatter plot for baseline generalized zero-shot action recognition framework~\cite{Xian2018}. On the right: t-SNE scatter plot for our approach based on an OD detector. Action categories are grouped into seen and unseen classes for illustration. The baseline GZSL~\cite{Xian2018} incorrectly classifies several unseen category features (denoted by 'FN Unseen') into seen action categories. Our approach significantly reduces the bias towards seen categories, resulting in accurate action recognition. Best viewed in color.  }
    \label{fig_tsne_hmdb}
\end{figure}

Contrary to action recognition in videos, extensive research efforts have been dedicated to zero-shot image classification. Most earlier ZSL approaches are based on attribute mapping~\cite{Zeynep2013,Lampert2009}. On the other hand, a few recent works ~\cite{fu2014transductive, li2015semi} tackle the 
problem in a transductive manner, by assuming access to the full set of unlabelled testing data. This helps in decreasing the domain shift problem, in ZSL, caused due to disjoint categories in training and testing. Similar transductive strategies have also been explored for action recognition in videos~\cite{Xu2017,wacv} to reduce the bias towards seen action categories. However, these approaches require unlabelled testing data for fine-tuning the parameters. Further, the bias still exists due to the similar treatment of both seen and unseen categories (see Fig.~\ref{fig_tsne_hmdb}(a)).
Instead, we propose a GZSL framework to separate the classification step  for the seen and unseen action classes by introducing an out-of-distribution (OD) detector. As a result, the inherently-learned bias towards the seen classes in the action classifier is  reduced (see Fig.~\ref{fig_tsne_hmdb}(b)).

In our approach, the out-of-distribution (OD) detector is learned to produce a non-uniform distribution with an emphasis (peaks) for seen categories and a uniformly distributed output for the unseen categories. This is achieived by utilizing an entropy loss to train our OD detector for maximizing the entropy of the output for unseen action category features. During inference, the entropy of the detector's output is compared to a specified threshold for determining whether the test feature belongs to a seen or unseen action category. Consequently, the test feature is dynamically routed to either of the two classifiers explicitly trained over seen and unseen classes, respectively, for final classification. Entropy loss has previously been used~\cite{springenberg15entropy} to train generative adversarial networks~\cite{GAN} (GAN) for image synthesis, in both unsupervised and semi-supervised settings. However, to the best of our knowledge, we are the first to propose the use of entropy loss in the construction of an OD detector for generalized zero-shot action recognition.

The proposed OD detector requires features from both seen and unseen action classes to avoid an assumption on the prior data distribution. However, unseen action features are not available during training. Thus, we propose to synthesize unseen action features, to train our OD detector, by adapting a conditional Wasserstein GAN~\cite{wgan} (WGAN) with additional terms: cosine embedding and cycle-consistency losses. The additional loss terms aid in improving the feature generation process for a diverse set of action categories. In our work, both the generator and discriminator of the WGAN  are conditioned on the category-specific auxiliary descriptions, called \emph{class-embeddings} or \emph{attributes}\footnote{Both these terms are used interchangeably in this work}, to synthesize class-specific action features. Consequently, our OD detector and the two action classifiers (seen and unseen) are trained using real and synthesized features from seen and unseen categories, respectively.

\noindent\textbf{Contributions:} We introduce a novel generalized zero-shot action recognition framework based on an out-of-distribution (OD) detector. Our OD detector is designed to reduce the effect of the inherent bias towards the seen action classes generally present in the standard GZSL framework. To synthesize unseen features for our OD detector training, we adapt the conditional Wasserstein GAN with additional loss terms. To the best of our knowledge, we are the first to introduce a GZSL action recognition framework based on an OD detector trained using real features from seen action categories and synthesized features from unseen action classes. Our OD detector efficiently discriminates the semantically similar seen and unseen action categories, leading to improved action classification. Our approach sets a new state-of-the-art for generalized zero-shot action recognition on three benchmarks.

\section{Related Work \label{sec_related}}
ZSL and GZSL have gained considerable attention in recent years since they can deal with challenging real-world problems, such as automatic tagging of images and videos with new categories previously unseen during training. Earlier approaches~\cite{Zeynep2013,Lampert2009,Lampert2014} for ZSL in images were based on direct or indirect attribute mapping between instances and their class attributes. 
Alternatively, several more recent works ~\cite{Norouzi2014,Changpinyo2016,Zeynep2016} determine the unseen classes based on the weighted combination of seen classes. 
In GZSL, obtaining realistic and discriminative training data for unseen classes to overcome the classifier's bias towards the seen classes is a challenge. Synthesizing visual features of unseen instances through an embedding-based matrix mapping to convert the ZSL problem to a typical supervised problem was explored in~\cite{Long-cvpr2017,Long-pami}. Approaches such as~\cite{Bucher2017,Xian2018,rafael2018} have used different variants of GANs~\cite{GAN} to generate synthetic unseen class features for the task of GZSL. Similar to~\cite{Xian2018,rafael2018}, we adapt the conditional WGAN~\cite{wgan} in our framework for generalized zero-shot action recognition.

In contrast to image classification, the problem of ZSL and GZSL for action recognition in videos has received less attention. Existing works pose the problem of ZSL and GZSL action recognition in the transductive setting, where unlabelled test data is also used during training~\cite{Xu2017, kodirov2015unsupervised, wacv}. A generative approach using Gaussians was used to synthesize unseen class data in~\cite{wacv}, where each action is represented as a probability distribution in the visual space. These works do not treat seen and unseen action classes separately, as proposed in this work. Further, these methods use unlabelled real features from the unseen classes to rectify the bias of the learned parameters towards the seen classes. Unlike these approaches, we do not use any unlabelled real features from unseen action classes in the training stage of our model. In~\cite{ijcai18}, action recognition under ZSL was addressed using a Fisher vector representation of traditional features and two-stream deep features with GloVE~\cite{glove} class embedding. However, the more challenging problem of GZSL action recognition was not addressed. A one-to-one comparison using different features, such as C3D~\cite{tran15c3d}, I3D~\cite{carreira17i3d}, also remains unexplored in the context of GZSL in these approaches.

\begin{figure*}[t]
    \centering
    \includegraphics[width=0.96\textwidth,height=0.31\textwidth]{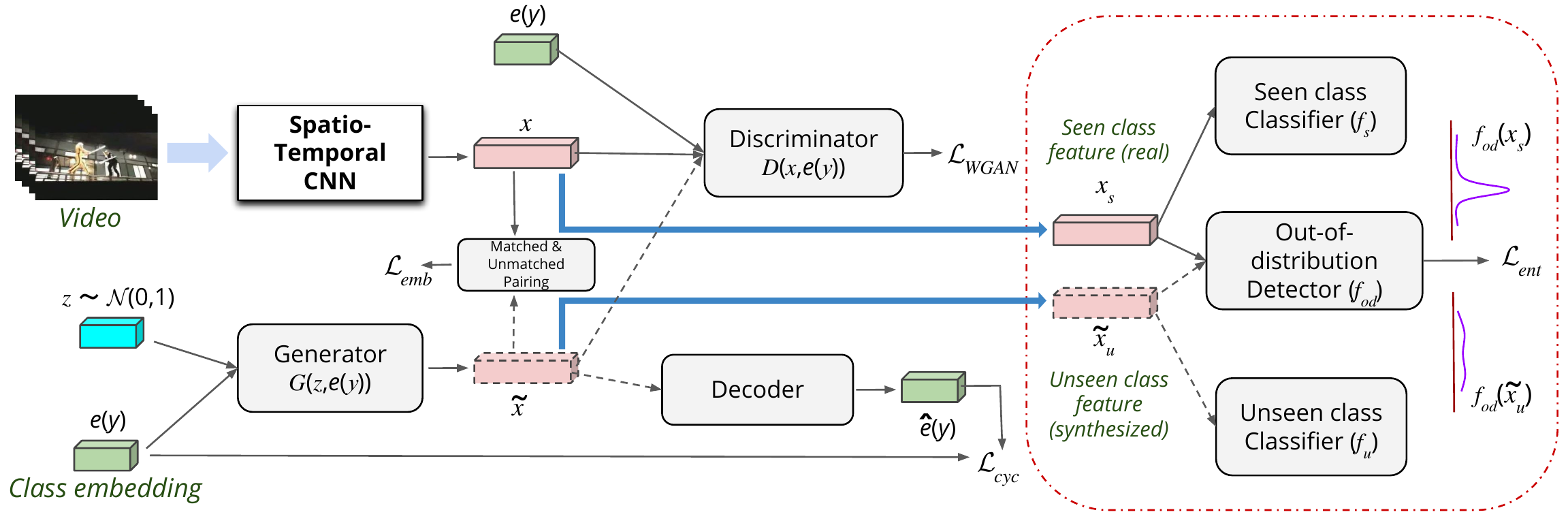}
    \caption{Illustration of the proposed GZSL approach: A conditional WGAN is trained to synthesize video features $\tilde{x}$, conditioned on the class embedding $e(y)$ via the losses $\mathcal{L}_{WGAN}$, $\mathcal{L}_{cyc}$ and $\mathcal{L}_{emb}$. A spatio-temporal CNN computes the real features $x$ for the seen class videos. During post-training, the generator, conditioned on the unseen class embedding $e(u)$, synthesizes unseen class features $\tilde{x}_u$, which, along with real features $x_s$, are used to learn the three classifiers $f_{od}$, $f_s$ and $f_u$. The expected outputs of $f_{od}$ for seen and unseen class features are also portrayed. Cuboids with dashed borders denote synthesized features. Dashed arrows indicate their corresponding path.
    }\vspace{-0.2cm}
    \label{fig_overall_pipeline}
\end{figure*}

Out-of-distribution detectors~\cite{lee18confidence-OD,OD18confidence} have been investigated in the context of image classification via cross-dataset evaluation. In~\cite{lee18confidence-OD}, instances that appear to be at the boundary of the data manifold were used as out-of-distribution examples during training while~\cite{OD18confidence} used the misclassified in-distribution samples as a proxy for out-of-distribution samples to calibrate the detector. However, in our approach, no such prior data distribution assumptions are made. Further, these detectors~\cite{lee18confidence-OD,OD18confidence} consider in-distribution samples from one image classification dataset and out-of-distribution samples from a different dataset, while our detector aims to distinguish between the seen and unseen class features of the same dataset.

\textbf{Our approach}: Different to the aforementioned works, an out-of-distribution detector is trained, with entropy loss, using GAN generated features of unseen action categories (as out-of-distribution samples) to recognize whether a feature sample belongs to either the seen or unseen group. Our method assumes no prior data distribution of the seen and unseen categories. The GAN itself is trained using the real features of seen categories, conditioned on the associated class-attributes of seen classes. During inference, based on the out-of-distribution detector's decision, features from a test instance are input to one of the two classifiers explicitly trained over seen and unseen action categories, respectively.

\section{Proposed Approach \label{sec_approach}}
The proposed framework for GZSL is detailed in this section. The framework is divided into two parts: synthetic video feature generation for unseen classes using GANs (Sec.~\ref{sec_synth_feat}) and out-of-distribution (OD) classifier learning (Sec.~\ref{sec_novelty_det}). The illustration of the overall pipeline is shown in Fig.~\ref{fig_overall_pipeline}. 

Let $\mathcal{S} = \{(x,y,e(y)| x \in \mathcal{X}, y \in \mathcal{Y}^s, e(y) \in \mathcal{E}\}$ be the training set for seen classes, where $x \in \mathbb{R}^{d_x}$ denotes the spatio-temporal CNN features, $y$ denotes the class labels in $\mathcal{Y}^s = \{y_1,\ldots,y_S\}$ with $S$ seen classes and $e(y) \in \mathbb{R}^{d_e}$ denotes the category-specific embedding that models the semantic relationship between the classes. Additionally, $\mathcal{U} = \{(u,e(u)| u \in \mathcal{Y}^u, e(u) \in \mathcal{E}\}$ is available during training, where $u$ is a class from a disjoint label set $\mathcal{Y}^u = \{u_1,\ldots,u_U\}$ of $U$ labels, and the corresponding videos or features are not available. The task in GZSL is to learn a classifier $f_{gzsl} : \mathcal{X} \rightarrow \mathcal{Y}^s \cup \mathcal{Y}^u $. Using the OD detector, this task can be reformulated into learning $3$ classifiers:  the out-of-distribution classifier $f_{od} : \mathcal{X} \rightarrow \{0,1\}$  and the seen and unseen classifiers $f_{s} : \mathcal{X} \rightarrow \mathcal{Y}^s $ and  $f_{u} : \mathcal{X} \rightarrow \mathcal{Y}^u $, respectively. The classifier $f_{od}$ will determine if the feature is an in-distribution or out-of-distribution feature and route it to either $f_s$ or $f_u$ to determine the class.

\subsection{Generating unseen class features\label{sec_synth_feat}}
Given the training data of seen classes, $\mathcal{S}$, the goal is to synthesize features belonging to unseen classes, $\tilde{x}$, using the class attributes, $e(u)$. To this end, a generative adversarial network (GAN) is learned using the seen class features, $x$, and the corresponding class embedding, $e(y)$. A GAN~\cite{GAN} consists of a generator $G$ and a discriminator $D$, which compete against each other in a two player minimax game. In the context of generating video features, $D$ attempts to accurately distinguish real-video features from synthetically generated features, while $G$ attempts to fool the discriminator by generating video features that are semantically close to real features. Since we need to synthesize features specific to unseen action categories, we use the conditional GAN~\cite{conditionalGAN} by conditioning both $G$ and $D$ on the embedding, $e(y)$. 
A conditional generator $G:\mathcal{Z}\times\mathcal{E} \rightarrow \mathcal{X}$ takes a random Gaussian noise $z \in \mathcal{Z}$ and a class embedding $e(y) \in \mathcal{E}$. Once the generator is learned, it is used to synthesize the video features of unseen classes, $u$, by conditioning on the unseen class embedding, $e(u)$. Further, we use the Wasserstein GAN~\cite{wgan} for the proposed framework due to its more stable training and recent success in~\cite{Xian2018,rafael2018} for zero-shot image classification.

A conditional WGAN~\cite{wgan}, conditioned on the embedding $e(y)$, is learned to synthesize the video features $\tilde{x}$, given the corresponding class embedding, $e(u)$. The conditional WGAN loss is given by 
\begin{equation}
    \mathcal{L}_{WGAN} = \mathbb{E}[D(x,e(y))] - \mathbb{E}[D(\tilde{x},e(y))] -  
\label{eqn_lwgan}
\end{equation}
\vspace{-0.5cm}
\begin{equation*}
    \quad \quad \quad \alpha\mathbb{E}[(||\nabla_{\hat{x}}D(\hat{x},e(y))||_2 - 1)^2]
\end{equation*}
where $\tilde{x}=G(z,e(y))$, $\hat{x}$ is a convex combination of $x$ and $\tilde{x}$, $\alpha$ is the penalty coefficient and $\mathbb{E}$ is the expectation. The first two terms approximate the Wasserstein distance in equation~\ref{eqn_lwgan}, with the third term being the penalty for constraining the gradient of $D$ to have unit norm along the convex combination of real and generated pairs. Additionally, we expect the generated features to be sufficiently discriminative such that the class embedding that generated them can be reconstructed back using the same features~\cite{CycleGAN}. To this end, similar to~\cite{rafael2018}, a decoder is used to reconstruct the class embedding $e(y)$ from the synthesized features $\tilde{x}$. Hence, a cycle-consistency loss is added to the loss formulation, which is given by,
\begin{equation}
\mathcal{L}_{cyc}=\mathbb{E}[||\hat{e}(y) -e(y)||_2]
\end{equation}
where $\hat{e}(y)$ is the reconstructed embedding.
Further, the synthesized features of a particular class $y_i$ should be similar to the real features of the same class and dissimilar to the features of other classes $y_j$ (for $j\neq i$). To this end, we first pair the real and synthesized features in a mini-batch to generate matched (same classes) and unmatched (different classes) pairs. Then, we minimize and maximize the distance between the matched and unmatched features, respectively, using the cosine embedding loss, as given by,
\begin{equation}
\mathcal{L}_{emb} = \mathbb{E}_{m}[1-cos(x,\tilde{x})] + \mathbb{E}_{um}[\max(0,cos(x,\tilde{x}))]
\vspace{-0.05cm}
\end{equation}
where the respective expectations are over the matched ($m$) and unmatched ($um$) pair distributions.
While the other losses ($\mathcal{L}_{WGAN}$ and $\mathcal{L}_{cyc}$) train the network by emphasizing the similarity between real and generated features of a particular class, the embedding loss also trains the network by emphasizing how the generated features of an action class should be dissimilar to the other class features. The final objective for training the GAN, using $\beta$ and $\gamma$ as hyper-parameters for weighting the respective losses,  is given by 
\begin{equation}
    \min_{G} \max_{D} \mathcal{L}_{WGAN} + \beta \mathcal{L}_{cyc} + \gamma \mathcal{L}_{emb} 
\label{eqn_final_obj}
\end{equation}

\subsection{Out-of-distribution detector for unseen class\label{sec_novelty_det}}
An out-of-distribution detector is proposed for differentiating between the features belonging to the seen classes and those belonging to unseen classes. After training the GAN using the training data $\mathcal{S}$, the generator ($G$) is used to synthesize features, $\tilde{x} = G(z,e(u))$, for the unseen categories $u \in \mathcal{Y}^u$. A training set of generated features, $\tilde{\mathcal{U}} = \{(\tilde{x},u,e(u))\}$, is obtained by generating sufficient features for all the unseen action categories. 

The real features of the seen classes, $x_s$ and the generated features of the unseen classes, $\tilde{x}_u$, are used to train the out-of-distribution detector. Approaches in~\cite{lee18confidence-OD,OD18confidence} learn an OD detector with a prior data distribution assumption of the seen class features. However, using generated samples of the unseen classes can help to better learn the boundaries between the seen and unseen categories, without assuming any prior data distribution. The detector is a fully-connected network with the dimension of the output layer the same as the number of seen classes, $S$. As shown in Sec.~\ref{sec_baseline_compare}, a binary classifier is insufficient to learn this task due to the complex boundaries between the many seen and unseen classes. Instead of attempting to  directly predict whether the input is from a seen or unseen class,
we use the concept of entropy to learn an embedding that projects the features of the seen and unseen classes far apart in the \emph{entropy space}. 
The network is trained with {entropy loss}, $\mathcal{L}_{ent}$, as given by

\begin{equation}
\mathcal{L}_{ent} = \mathbb{E}_{x\sim\mathcal{S}}[H(p_s)] -\mathbb{E}_{\tilde{x}\sim\mathcal{U}}[H(\tilde{p}_u)] 
\label{eqn_entropy_loss}
\end{equation}
where  
$H(p)$=$-\sum_i p[i]\log(p[i])$ is the entropy of $p$, 
and $p_s$=$f_{od}(x_s)$ and $\tilde{p}_u$=$f_{od}(\tilde{x}_u) \in \mathbb{R}^S$ are the predictions of the network for the seen and unseen features $x_s$ and $\tilde{x}_u$, respectively. 
Further, a negative log-likelihood term $N(p_s)$=$-\log(p_s[y_s])$, where $y_s$ is the class label of $x_s$, is added to Eq.~\ref{eqn_entropy_loss} for faster convergence. 
This type of loss formulation models the output of the network such that its entropy is minimum and maximum for the input features of seen and unseen classes, respectively. The higher the entropy, the higher the uncertainty. Thus, a seen class feature input will have a non-uniformly distributed output (with an emphasis on seen classes).
Similarly, an unseen class feature will have a near-uniform distribution as its output. 
The expected output of the classifier, $f_{od}$, for the seen and unseen class features is illustrated in the far-right side of Fig.~\ref{fig_overall_pipeline}. 

\textbf{Seen and unseen classifiers}: Alongside the OD detector training, we also train two separate classifiers, one for the seen classes and one for the unseen classes. The two classifiers $f_s$ and $f_u$ are trained on real features of seen classes $x_s$ and generated features of unseen classes $\tilde{x}_u$, respectively. During inference, the test video is passed through a spatio-temporal CNN to compute the real features $x_{test}$ and then sent to the OD detector. If the entropy of the output $f_{od}(x_{test})$ is less than a threshold $ent_{th}$, the feature $x_{test}$ is passed through the seen-classes classifier $f_s$ in order to predict the label of the test video. If the entropy of $f_{od}(x_{test})$ is greater than $ent_{th}$, then the label is predicted using the unseen-classes classifier $f_u$.  
In ZSL, where the test samples are restricted to belonging to unseen classes, only the unseen-classes classifier $f_u$ is required to predict the category of the video. In summary, the OD detector separates the classification of seen and unseen categories and reduces the bias towards seen categories.

\section{Experiments\label{sec_exp}}
\subsection{Experimental setup \label{sec_exp_setup}}
\noindent\textbf{Video features}:
Two types of video features, I3D~\cite{carreira17i3d} (Inflated 3D) and C3D~\cite{tran15c3d} (Convolution 3D), designed for generic action recognition, are used for evaluation. 
The appearance and flow I3D features are extracted from the \emph{Mixed{\_}5c} layer output of the RGB and flow I3D networks, respectively. Both networks are pretrained on the Kinetics dataset~\cite{carreira17i3d}. For an input video, the \emph{Mixed{\_}5c} output of both networks are averaged across the temporal dimension and pooled by 4 in the spatial dimension and then flattened to obtain a vector, of size 4096, representing the appearance and flow features, respectively. The appearance and flow features are concatenated to obtain video features of size 8192.
We use the C3D model, pre-trained on the Sports-1M dataset~\cite{karpathy14sports1m}, to extract the C3D features for representing the actions in a video. A video is divided into non-overlapping 16-frame clips and the mean of the \emph{fc6} layer outputs, of size 4096, is taken as the video feature for the action. 

\noindent\textbf{Network architecture}:
The generator $G$ is a three-layer fully-connected (FC) network with an output layer dimension equal to the size of the video feature. The hidden layers are of size 4096. The decoder is also a three-layer FC network with an output size equal to the class-embedding size and a hidden size equal to 4096. The discriminator $D$ is a two-layer FC network with the output size equal to 1 and a hidden size equal to 4096. The individual classifiers $f_s$ and $f_u$ are single-layer FC networks with an input size equal to the video feature size and output sizes equal to the number of seen and unseen classes, respectively. The OD detector $f_{od}$ is a three-layer FC network with output and hidden layer sizes equal to the number of seen classes and 512, respectively. The parameters $\beta$ and $\gamma$ are set to 0.01 and 0.1, respectively, for all the datasets. The threshold value $ent_{th}$ is chosen to be the average of the prediction entropies of the seen class features in the training data. All the modules are trained using the Adam optimizer with a 10${}^{-4}$ learning rate.

\noindent\textbf{Datasets}:
Three challenging video action datasets (Olympic Sports~\cite{olympic}, HMDB51~\cite{hmdb} and UCF101~\cite{ucf-dataset}), widely used as benchmarks for GZSL and ZSL, are used for evaluating the performance of the proposed technique. The details of the three datasets are given in Tab.~\ref{tab_dataset}. The mean per-class accuracy averaged over 30 independent test runs is reported along with the standard deviation. Each test run is carried out on a random split of the seen and unseen classes in the dataset. For GZSL, we also report the mean accuracy for the seen classes, mean accuracy of the unseen classes and the harmonic mean of the two. For the GZSL setting, the test data consists of all the videos belonging to unseen classes and a random subset of 20\% videos from seen class categories. 
\begin{table}[t]
\centering
\adjustbox{width=\linewidth}{ 
\begin{tabular}{|l|c|c|c|}
\hline
\textbf{Dataset} & \textbf{\#Videos} & \textbf{\#Class} & \textbf{Split (Seen / Unseen)} \\ \hline
\textbf{Olympic Sports} & 783 & 16 & 8/8 \\ \hline
\textbf{HMDB51} & 6766 & 51 & 26/25 \\ \hline
\textbf{UCF101} & 13320 & 101 & 51/50 \\ \hline
\end{tabular}
}
\caption{\label{tab_dataset}Datasets used for evaluation}\vspace{-0.2cm}
\end{table}

\noindent{\bf Class-embedding}: We use two types of class-embedding to semantically represent the classes: the human annotated attributes and \emph{word vectors}~\cite{mikolov13w2v}. The UCF101 and Olympic Sports datasets also have manually-annotated class attributes of sizes 40 and 115, respectively. A skip-gram model, trained on the news text corpus provided by Google, is used to generate the action class-specific word vector representations of size 300 using the action category names as input. 
The HMDB51 dataset does not have any associated manual attributes.

\begin{table*}[t]
\centering
\begin{tabular}{|l|l|c|r|r|r|r|r|r|r|r|r|}
\hline
 & & \textbf{Embed} & \multicolumn{3}{c|}{\bf Olympic Sports} & \multicolumn{3}{c|}{\bf HMDB51} & \multicolumn{3}{c|}{\bf UCF101} \\ 
 \hline \hline
 & & \multicolumn{1}{l|}{} & \multicolumn{1}{c|}{$s$} & \multicolumn{1}{c|}{$u$} & \multicolumn{1}{c|}{$H$} & \multicolumn{1}{c|}{$s$} & \multicolumn{1}{c|}{$u$} & \multicolumn{1}{c|}{$H$} & \multicolumn{1}{c|}{$s$} & \multicolumn{1}{c|}{$u$} & \multicolumn{1}{c|}{$H$} \\ \hline
\multirow{2}{*}{(a)} & \multirow{2}{*}{f-CLSWGAN${}^*$~\cite{Xian2018}} & w2v & 66.0 & 35.5 & 46.1 & 52.6 & 23.7 & 32.7 & 74.8 & 20.7 & 32.4 \\ \cline{3-12} 
& & manual & 72.3 & 51.1 & 59.9 & - & - & - & 83.9  & 30.2  & 44.4 \\ 
 \hline
 \hline
\multirow{2}{*}{(b)} & {$\mathcal{L}_{WGAN}+\mathcal{L}_{cyc}+\mathcal{L}_{emb}$} & w2v & 67.6 & 36.5 & 47.4 & 51.7 & 24.9 & 33.6 & 73.7 & 21.8 & 33.7 \\ \cline{3-12} 
& (\textbf{Ours: CEWGAN}) & manual & 73.7 & 52.3 & 61.1 & - & - & - & 80.2 & 31.7 & 45.5 \\ 
 \hline
 \hline
\multirow{2}{*}{(c)} & \multirow{2}{*}{(b) + OD$_{bin}$} & w2v & 71.6 & 38.1 & 49.8 & 36.7 & 24.1 & 29.1 &  62.4  & 19.2  & 29.4 \\ \cline{3-12} 
& & manual & 72.1 & 56.9 & 63.6 & - & - & - & 67.4  & 28.2  & 39.8 \\ 
 \hline
 \hline
\multirow{2}{*}{\textbf{(d)}} & {(b) + OD$_{ent}$} & w2v & 73.2 & 41.8 & \bf{53.1} & 55.6 & 26.8 & \bf{36.1} & 75.9 & 24.8 & \bf{37.3} \\ \cline{3-12} 
& (\textbf{Ours: CEWGAN-OD}) & manual & 71.5 & 61.6 & \bf{66.2} & - & - & - & 76.7 & 36.4 & \bf{49.4} \\ 
 \hline
 
\end{tabular}
\vspace{0.02cm}
\caption{\label{tab_baseline_compare}Comparison of proposed approach with the baseline f-CLSWGAN${}^*$~\cite{Xian2018} (* - adapted implementation) using concatenated I3D features for GZSL action recognition. CEWGAN-OD and CEWGAN denote the proposed framework with and without the out-of-distribution (OD) detector, respectively. OD$_{bin}$ and OD$_{ent}$ denote the binary classifier and proposed OD detectors, respectively. Higher is better. Manual attributes are not available for HMDB51. $s$, $u$ and $H$ denote the accuracy for seen and unseen classes and their harmonic mean, respectively. CEWGAN outperforms the baseline f-CLSWGAN on all datasets. Integrating OD$_{ent}$ with CEWGAN achieves further gains. }\vspace{-0.2cm}
\end{table*}

\subsection{Baseline comparison \label{sec_baseline_compare}}
The proposed framework is compared with the baseline by evaluating on the generalized zero-shot action recognition task using I3D concatenated features. Since our GAN framework for synthesizing features also uses the WGAN~\cite{wgan}, we choose f-CLSWGAN~\cite{Xian2018}, originally designed for zero-shot image classification, as the baseline. The performance comparison for the three datasets is shown in Tab.~\ref{tab_baseline_compare}. We also compare our GZSL framework with and without the OD detector (denoted as CEWGAN-OD and CEWGAN, respectively, in Tab.~\ref{tab_baseline_compare}). Further, to quantify the effectiveness of our OD detector, we also combine CEWGAN with a binary OD classifier, OD${}_{bin}$. The classification accuracy for the seen and unseen categories and their harmonic mean are denoted by $s$, $u$ and $H$ , respectively.

The proposed OD detector (OD${}_{ent}$) always outperforms the binary OD detector (OD${}_{bin}$) (see Tab.~\ref{tab_baseline_compare}), proving that a binary classifier is not sufficient for learning the task. The OD${}_{bin}$ requires generated features for seen and unseen classes to achieve reasonable performance and it still fares, generally, worse than CEWGAN. It only yields better results than CEWGAN for the Olympic Sports dataset. The main reason is that Olympic Sports has only eight seen and unseen classes. Hence, it is easier to separate the corresponding test features. As the number of classes increases, OD${}_{bin}$ fails to accurately separate the seen and unseen category features.

Importantly, we see that the proposed GAN (CEWGAN) performs better than the baseline approach (f-CLSWGAN) on all combinations of datasets and attributes. Integrating the proposed OD detector (OD${}_{ent}$) with CEWGAN further improves the performance across datasets. Average gains of 7.0\%, 3.4\%, and 4.9\% (in terms of accuracy) are achieved over f-CLSWGAN~\cite{Xian2018} for the Olympic Sports, HMDB51 and UCF101 datasets, respectively, using \emph{word2vec}. Achieving a considerable gain on a difficult dataset, such as HMDB51, shows the promise of our framework for generalized zero-shot action recognition. 

\subsection{State-of-the-art comparison \label{sec_sota_compare}}
In this section, a comparison of our proposed framework against the other approaches for the tasks of ZSL and GZSL in action recognition is given. Since our aim is reducing the bias of the classifier towards seen classes in generalized zero-shot action recognition, we first compare the GZSL performance (Tab.~\ref{tab_gzsl_sota}), and then the ZSL performance (Tab.~\ref{tab_zsl_sota}), with the other approaches in literature. In both the tables, we report the performance of our approach trained using the I3D (appearance + flow) features. The performance of our approach using other features is given as an ablation study in Sec.~\ref{sec_feat_compare}.

\textbf{GZSL performance comparison}: The proposed out-of-distribution detector is applicable only in the GZSL framework. 
The comparison of our proposed approach with the other approaches on the GZSL task is reported in Tab.~\ref{tab_gzsl_sota}. The best results for each dataset and attribute combination are in boldface. The standard deviation from the mean is also reported. We see that the proposed approach, CEWGAN-OD, outperforms the other approaches (fewer approaches compared to the ZSL task) on all datasets. The results for CLSWGAN~\cite{Xian2018} are obtained by adapting the author's implementation for our GZSL action recognition task. This is denoted by the superscript '*' in Tab.~\ref{tab_gzsl_sota}. Both CLSWGAN and the proposed approach are trained using the I3D features. The best existing approach for GZSL action recognition, GGM~\cite{wacv}, employs a generative approach to synthesize unseen class data and utilizes unlabelled real features (C3D) from the unseen classes to rectify the bias of the learned parameters towards seen classes. Particularly, for the UCF101 dataset and manual attributes combination, the proposed approach, CEWGAN-OD, achieves gains of 5.1\% and 25.8\% (in terms of accuracy) over the CLSWGAN~\cite{Xian2018} and GGM~\cite{wacv}, respectively. 
Further, for the \emph{word2vec} embedding, the proposed CEWGAN-OD achieves gains of 16\% and 19.8\% over the best existing approach, GGM~\cite{wacv}, for the HMDB51 and UCF101 datasets, respectively.

\begin{table}[t]
\small
\hspace{-0.1cm}
\adjustbox{width=1.01\linewidth}{
\begin{tabular}{|l|c|c|c|c|}
\hline
\multicolumn{1}{|c|}{\textbf{Method}} &  & \textbf{Olympics} & \textbf{HMDB51} & \textbf{UCF101} \\ \hline
HAA~\cite{liu11recognizing} & \emph{m} & 49.4\rpm 10.8 & - & 18.7\rpm 2.4 \\ \hline
SJE~\cite{akata15evaluation} & \emph{w} & 32.5\rpm 6.7 & 10.5\rpm 2.4 & 8.9\rpm 2.2 \\ \hline
ConSE~\cite{Norouzi2014} & \emph{w} & 37.6\rpm 9.9 & 15.4\rpm 2.8 & 12.7\rpm 2.2 \\ \hline
\multirow{2}{*}{GGM~\cite{wacv}} & \emph{m} & 52.4\rpm 12.2 & - & 23.7\rpm 1.2 \\ \cline{2-5} 
 & \emph{w} & 42.2\rpm 10.2 & 20.1\rpm 2.1 & 17.5\rpm 2.2 \\ \hline
{CLSWGAN$^*$} & \emph{m} & 59.9\rpm 5.3  & - & 44.4\rpm 3.0 \\ \cline{2-5} 
 ~\cite{Xian2018} & \emph{w} & 46.1\rpm 3.7  & 32.7\rpm 3.4  & 32.4\rpm 3.3 \\ \hline
 \hline
{\bf Ours:} & \emph{m} & \textbf{66.2\rpm 6.3} & - & \textbf{49.4\rpm 2.4} \\ \cline{2-5} 
{\bf CEWGAN-OD} & \emph{w} & \textbf{53.1\rpm 3.6} & \textbf{36.1\rpm 2.2} & \textbf{37.3\rpm 2.1} \\ \hline
\end{tabular}
}
\caption{\label{tab_gzsl_sota}GZSL performance comparison (in \%) with existing approaches. \emph{m} and \emph{w} indicate manual attributes and \emph{word2vec}, respectively. CLSWGAN${}^*$~\cite{Xian2018} (* - adapted implementation) and CEWGAN-OD denote the baseline and proposed approach, respectively, using I3D features. Higher is better. Best results for each embedding are in bold. Manual attributes are not available for HMDB51. CEWGAN-OD achieves an absolute gain of 5.0\% over the baseline for UCF101, using manual attributes, and outperforms existing methods by a significant margin on all datasets.}\vspace{-0.1cm}
\end{table}

\begin{table}[t]
\centering
\adjustbox{width=\linewidth}{   
\small
\begin{tabular}{|l|c|c|c|c|}
\hline
\multicolumn{1}{|c|}{\textbf{Method}} &  & \textbf{Olympics} & \textbf{HMDB51} & \textbf{UCF101} \\ \hline
PST~\cite{rohrbach13transfer} & \emph{m} & 48.6\rpm 11 & - & 15.3\rpm 2.2 \\ \hline
ST~\cite{xu15semantic} & \emph{w} & - & 15\rpm 3 & 15.8\rpm 2.3 \\ \hline
\multirow{2}{*}{TZWE~\cite{Xu2017}} & \emph{m} & 53.5\rpm 11.9 & - & 20.2\rpm 2.2 \\ \cline{2-5} 
 & \emph{w} & 38.6 \rpm 10.6 & 19.1\rpm 3.8 & 18.0\rpm 2.7 \\ \hline
\multirow{2}{*}{Bi-dir~\cite{wang17bidir}} & \emph{m} & - & - & 28.3\rpm 1.0 \\ \cline{2-5} 
 & \emph{w} & - & 18.9\rpm 1.1 & 21.4\rpm 0.8 \\ \hline
UDA~\cite{kodirov2015unsupervised} & \emph{m} & - & - & 13.2\rpm 0.6 \\ \hline
\multirow{1}{*}{MICC}~\cite{ijcai18} & \emph{g} & 43.9\rpm 7.9 & 25.3\rpm 4.5 & 25.4\rpm 3.1 \\ \hline 
\multirow{2}{*}{GGM~\cite{wacv}} & \emph{m} & 57.9\rpm 14.1 & - & 24.5\rpm 2.9 \\ \cline{2-5} 
 & \emph{w} & 41.3\rpm 11.4 & 20.7\rpm 3.1 & 20.3\rpm 1.9 \\ \hline
{CLSWGAN$^*$} & \emph{m} & 64.7\rpm 7.5  & - & 37.5\rpm 3.1  \\ \cline{2-5} 
 ~\cite{Xian2018}& \emph{w} & 47.1\rpm 6.4  & 29.1\rpm 3.8 & 25.8\rpm 3.2  \\ \hline
 \hline
{\bf Ours:} & \emph{m} & \textbf{65.9\rpm 8.1} & -  & \textbf{38.3\rpm 3.0} \\ \cline{2-5} 
{\bf CEWGAN}& \emph{w} & \textbf{50.5\rpm 6.9} & \textbf{30.2\rpm 2.7} & \textbf{26.9\rpm 2.8} \\ \hline
\end{tabular}
}
\caption{\label{tab_zsl_sota}ZSL performance comparison (in \%) with existing approaches. \emph{m}, \emph{g} and \emph{w} indicate manual attributes, \emph{GLoVE} and \emph{word2vec}, respectively. CLSWGAN${}^*$~\cite{Xian2018} (* - adapted implementation) and CEWGAN denote the baseline and proposed approach, respectively, using I3D features. Higher is better. Best results for each embedding are in bold. Our approach achieves the state-of-the-art on all datasets. }\vspace{-0.2cm}
\end{table}

\textbf{ZSL performance comparison}: In Tab.~\ref{tab_zsl_sota}, the proposed approach trained using the I3D (appearance + flow) features is denoted by CEWGAN. Here, the suffix OD (used in Tab.~\ref{tab_gzsl_sota}) is dropped since the out-of-distribution detector is applicable only in the GZSL task. 
From Tab.~\ref{tab_zsl_sota}, we see that our approach outperforms the other approaches in the zero-shot action recognition task for all combinations of datasets and attributes. The proposed approach, CEWGAN, in general, has less or comparable deviation as the other approaches. This shows that the proposed approach consistently improves across the splits.
All the other approaches use either the \emph{word2vec} or manually-annotated embedding (denoted by \emph{w} and \emph{m}, respectively) except MICC~\cite{ijcai18}, which uses \emph{GloVE}~\cite{glove}, an embedding similar to \emph{word2vec}. The proposed approach using I3D features and the \emph{word2vec} embedding has absolute gains of $6.6\%$, $4.9\%$ and $1.5\%$ (in terms of accuracy) over the best existing ZSL results on the Olympic Sports, HMDB51 and UCF101 datasets, respectively. Further, for the \emph{word2vec} embedding, we observe that the proposed CEWGAN achieves gains of 1.2\%, 1.1\% and 1.1\% over the CLSWGAN~\cite{Xian2018} for the same datasets, respectively. Generally, for both GZSL and ZSL tasks, using the same features but learning with manual attributes (instead of \emph{word2vec}) results in better performance across different approaches.

\subsection{Bias towards seen categories}
Tab.~\ref{tab_seen_bias} quantifies the bias reduction due to the proposed framework, CEWGAN-OD, for the three datasets, using the \emph{word2vec} embedding. For this experiment, we consider all the features of unseen categories as one class and the remaining features from seen categories as another. A feature sample is said to be wrongly classified if the predicted class is not the same as the ground-truth class, regardless of whether the feature was classified as belonging to the correct category within each class or not. This allows us to quantify the bias reduction achieved by the standalone OD detector. We observe that CEWGAN-OD reduces the bias towards the seen categories and achieves better classification for the unseen class features. Specifically, the proposed CEWGAN-OD achieves gains of 6.2\% and 10.1\% over CEWGAN for the HMDB51 and UCF101 datasets, respectively, using the \emph{word2vec} embedding. 

\begin{figure}[t]
    \centering
    \includegraphics[width=0.43\textwidth,height=0.32\textwidth]{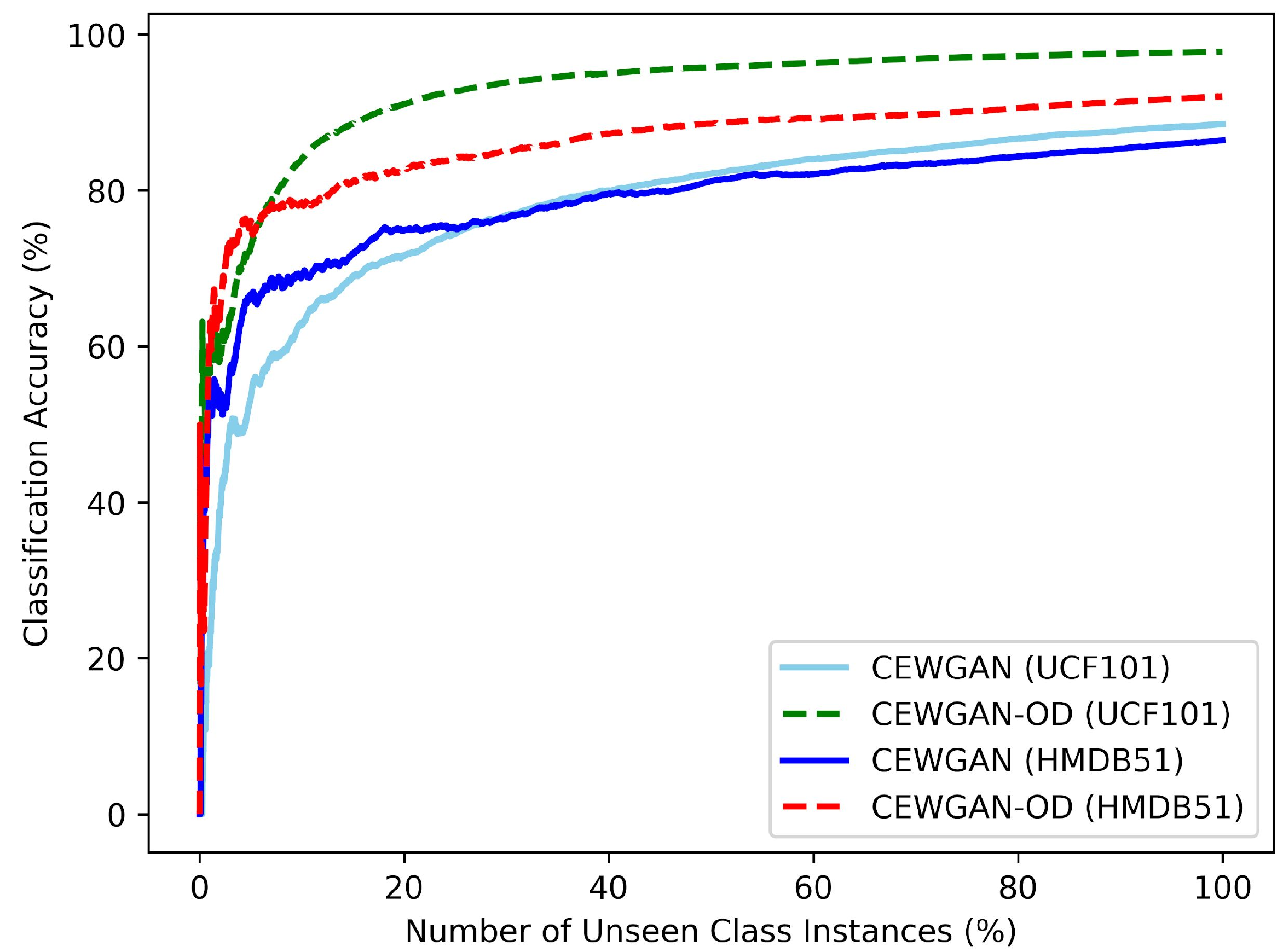}
    \caption{Classification accuracy (in \%) comparison between the proposed GZSL frameworks (CEWGAN-OD and CEWGAN) for random test splits of HMDB51 and UCF101 datasets. X-axis denotes the number of unseen class instances (in \%). For each framework, the unseen feature instances are sorted (in descending order) according to their respective classifier scores. Integrating the OD detector in the CEWGAN framework achieves higher classification accuracy (red and green lines) for both datasets. CEWGAN-OD decreases the bias towards seen classes. Best viewed in color.}\vspace{-0.22cm}
    \label{fig_acc_plot}
\end{figure}

Fig.~\ref{fig_acc_plot} shows a comparison, in terms of the classification accuracy, between our two
frameworks: CEWGAN and CEWGAN-OD. The comparison is shown for random test splits of HMDB51 and UCF101. The x-axis denotes the number of unseen class feature instances in a test split. The unseen class feature instances are sorted (high to low) according to the confidence scores of the respective classifiers (CEWGAN and CEWGAN-OD). The plot shows that integrating the proposed OD detector in the GZSL framework results in a significant improvement in performance for both datasets (denoted by green and red curves in Fig.~\ref{fig_acc_plot}). The number of unseen class feature instances incorrectly classified (into a seen class) is reduced with the integration of the proposed OD dectector. This improvement in classification performance for unseen action categories leads to a significant reduction in bias towards seen classes.

\begin{table}[t]
\centering
\adjustbox{width=0.8\linewidth}{   
\small
\begin{tabular}{|l|c|c||c|c|}
\hline
\textbf{} & \multicolumn{2}{c||}{\textbf{CEWGAN}} & \multicolumn{2}{c|}{\textbf{CEWGAN-OD}} \\ \hline
 & SC & UC & SC & UC \\ \hline
\textbf{Olympic Sports} & 68.2  & 72.3 & \textbf{73.9} & \textbf{82.8}  \\ \hline
\textbf{HMDB51} & 66.7 & 82.5 & \textbf{71.6} & \textbf{88.7} \\ \hline
\textbf{UCF101} & 74.4 & 81.1 & \textbf{76.5} & \textbf{92.2} \\ \hline
\end{tabular}
}
\vspace{0.02cm}
\caption{\label{tab_seen_bias}Comparison of the bias towards seen classes, between the baseline (CEWGAN) and the proposed (CEWGAN-OD) frameworks on the three datasets using the \emph{word2vec} embedding. SC, UC denote seen classes and unseen classes, respectively. Lower UC accuracy indicates higher bias towards seen categories. The proposed CEWGAN-OD achieves gains of 6.2\% and 10.1\% (classification accuracy) over the baseline CEWGAN for the unseen categories in the HMDB51 and UCF101 datasets, respectively.}\vspace{-0.22cm}
\end{table}

\subsection{Transferring word representations \label{sec_w2vec2manual}}
As mentioned previously in Sec.~\ref{sec_exp_setup}, manual attributes are not available for the HMDB51 dataset. While \emph{word2vec} representations give a good measure of the semantic representations of the classes, learning with manual attributes always results in better performance, as can be seen from the results in Sec.~\ref{sec_sota_compare}~and~\ref{sec_baseline_compare}. Here, we learn to generate the manual attributes from the \emph{word2vec} embedding to show that using the transformed class embedding achieves better generation of features, resulting in better performance compared to the \emph{word2vec} embedding. We use the class embeddings of the UCF101 dataset to learn the transformation using a two-layer FC network. 
To generate a sufficient number of samples for training, the video features are concatenated with their respective \emph{word2vec} and used as input. The trained model is then used to transform \emph{word2vec} representations into manual attribute embeddings.

To comply with the ZSL paradigm of not using any video features from the unseen classes, we use the generated features for the HMDB51 unseen classes as input for the embedding transformation network. Here, the generator is learned using the \emph{word2vec} embedding and the seen class features of the HMDB51 dataset. The learned attributes for HMDB51 are the same size as the manual attributes of UCF101, \ie, $115$. The performance of the proposed framework under ZSL and GZSL settings for the HMDB51 dataset using the transferred attributes (denoted by \emph{m}) and different features is reported in Tab.~\ref{tab_feat_compare}.
The results show that the transferred attributes for HMDB51 achieve better performance than the \emph{word2vec}. Hence, synthesizing features using transferred attributes, for datasets without manually-annotated attributes, achieves better performance compared to synthesizing using the standard \emph{word2vec} embedding.

\subsection{Comparison of video features \label{sec_feat_compare}}
Here, we give a performance comparison of the different video features for the tasks of ZSL and GZSL. The features that are used for comparison are C3D, I3D$_{a}$ (appearance), I3D$_{f}$ (flow) and I3D$_{af}$ (appearance and flow). The features are evaluated on the HMDB51 and UCF101 datasets using both the manual attributes and \emph{word2vec} embedding. The manual attributes for HMDB51 refer to the transformed attributes, as described in Sec.~\ref{sec_w2vec2manual}. The entire setup remains the same except for the input or output layers, which depend on the video feature dimensions. The results are reported in Tab.~\ref{tab_feat_compare}. In general, we see that the I3D$_{a}$ features perform better than the C3D and I3D$_{f}$ features. The I3D$_{f}$ features are still better than the C3D features, while the best performance is achieved when the appearance and flow features are combined. This is in line with the performance of the features in the task of fully-supervised action recognition, as noted in~\cite{carreira17i3d}. This also indicates that our framework can be used with new and improved features as and when they are designed and a corresponding improvement in GZSL action recognition can be expected.
The results in Tab.~\ref{tab_gzsl_sota}~and~\ref{tab_zsl_sota} for CEWGAN-OD and CEWGAN, respectively, use the combined features, I3D$_{af}$. 
\begin{table}[t]
\centering
\small
\adjustbox{width=0.8\linewidth}{
\begin{tabular}{|l|c|c|c|c|c|}
\hline
\multicolumn{1}{|c|}{\textbf{Feature}} &  & \multicolumn{2}{c|}{\textbf{HMDB51}} & \multicolumn{2}{c|}{\textbf{UCF101}} \\ \hline
 &  & {ZSL} & {GZSL} & {ZSL} & {GZSL} \\ \hline
C3D & \emph{m} & 26.0  & 30.9 & 28.1 & 38.7 \\ \hline
 & \emph{w} & 24.2 & 29.1 & 21.5 & 32.0 \\ \hline
I3D$_a$ & \emph{m} & 30.8 & 36.1 & 33.9 & 44.3 \\ \hline
 & \emph{w} & 28.2 & 33.8 & 23.2 & 33.4 \\ \hline
I3D$_f$ & \emph{m} & 29.7 & 34.9 & 32.2 & 42.7 \\ \hline
 & \emph{w} & 27.4  & 32.0 & 22.7 & 32.6 \\ \hline
I3D$_{af}$ & \emph{m} & \textbf{34.8} & \textbf{39.5} & \textbf{38.3} & \textbf{49.4}   \\ \hline
 & \emph{w} & \textbf{30.2} & \textbf{36.1} & \textbf{26.9} & \textbf{37.3}  \\ \hline
\end{tabular}
}
\vspace{0.02cm}
\caption{\label{tab_feat_compare}Performance comparison of C3D, I3D appearance (I3D$_{a}$), I3D flow (I3D$_{f}$) and I3D appearance+flow (I3D$_{af}$) video features on the HMDB51 and UCF101 datasets. For HMDB51, \emph{m} denotes the transferred attributes, as discussed in Sec.~\ref{sec_w2vec2manual}. Best results are in bold for both types of embedding. For every combination of feature and attribute, ZSL and GZSL denote the performance of CEWGAN and CEWGAN-OD, respectively.}\vspace{-0.2cm}
\end{table}

\section{Conclusion \label{sec_conclusion}}
In this work, we proposed a novel out-of-distribution detector integrated into the generalized zero-shot learning action recognition framework. An out-of-distribution detector was learned to detect unseen category features as out-of-distribution samples. It was trained using real and GAN-generated features from seen and unseen categories, respectively. 
The use of an out-of-distribution detector enabled the classification of the seen and unseen categories to be separated and hence, reduced the bias towards seen classes that is present in the baseline approaches. The approach was evaluated on three human action video datasets, using different types of embedding and video features. The proposed approach outperformed the baseline~\cite{Xian2018} in generalized zero-shot action recognition using \emph{word2vec}, with absolute gains of $7.0\%$, $3.4\%$ and $4.9\%$ on the Olympic Sports, HMDB51 and UCF101 datasets, respectively.

\section*{Acknowledgement}
Part of the research was carried out when Devraj Mandal and Sanath Narayan were at Mercedes-Benz R\&D India.

{\small
\bibliographystyle{ieee_fullname}
\bibliography{main}
}
\end{document}